\documentclass[letterpaper, 10 pt, conference]{ieeeconf}  
\usepackage{graphicx}
\usepackage[table]{xcolor}
\usepackage{tcolorbox}
\usepackage{nicematrix,tikz}
\usepackage[table]{xcolor}
\usepackage{svg}
\usepackage{comment}
\usepackage{float}
\usepackage{acronym}
\usepackage{amsmath}
\usepackage{balance}
\usepackage{gensymb} 
\usepackage{units} 
\usepackage{bm}
\usepackage{tabularx}
\usepackage{array}
    \newcolumntype{P}[1]{>{\centering\arraybackslash}p{#1}}
    \newcolumntype{M}[1]{>{\centering\arraybackslash}m{#1}}
\usepackage{hyperref}
\graphicspath{ {./images/} }

\IEEEoverridecommandlockouts                              

\usepackage{amssymb}  

\usepackage{amsmath}
\usepackage{amssymb}

\newcommand\referencediag[4]{{}^{{#2}}_{{#4}}{{#1}}^{}_{{#3}}}   
\newcommand\referencediagt[5]{{}^{{#2}}_{{#4}}{{#1}}^{#5}_{{#3}}}   
\newcommand\reference[4]{{\referencediag{#1}{#2}{#3}{#4}}}  
\newcommand\referencet[5]{{\referencediagt{#1}{#2}{#3}{#4}{#5}}}  

\newcommand\bSigma{{\mathbf{\Sigma}}}

\makeatletter
\newcommand{\generateAlphabet}[4]{%
  \def\@tempa{#1} 
  \count@=`#3
  \loop
  \begingroup\lccode`?=\count@
  \lowercase{\endgroup\@namedef{\@tempa ?}{#2{?}}}%
  \ifnum\count@<`#4
  \advance\count@\@ne
  \repeat
}

\generateAlphabet{v}{\mathbf}{A}{Z}
\generateAlphabet{v}{\mathbf}{a}{z}
\generateAlphabet{c}{\mathcal}{A}{Z}
\generateAlphabet{cc}{\mathcal}{a}{z}
\generateAlphabet{cs}{\scriptscriptstyle\mathcal}{A}{Z}
\generateAlphabet{fr}{\mathfrak}{A}{Z}
\generateAlphabet{fr}{\mathfrak}{a}{z}
\generateAlphabet{rm}{\mathrm}{A}{Z}
\generateAlphabet{rm}{\mathrm}{a}{z}

\acrodef{FOV}{Field Of View}
\acrodef{RCS}{Radar Cross Section}
\acrodef{IF}{intermediate frequency}
\acrodef{RO}{Radar Odometry}
\acrodef{FFT}{Fast-Fourier Transform}
\acrodef{FMCW}{Frequency Modulated Continuous Wave}
\acrodef{ToF}{Time-of-Flight}
\acrodef{AoA}{Angle-of-Arrival}
\acrodef{ADC}{Analog-to-Digital Converter}
\acrodef{IMU}{Inertial Measurement Unit}
\acrodef{EKF}{Extended Kalman Filter}
\acrodef{VIO}{Visual-Inertial Odometry}
\acrodef{GNSS}{Global Navigation Satellite System}
\acrodef{UAV}{Unmanned Aerial Vehicle}
\acrodef{CNN}{Convolutional Neural Network}
\acrodef{RIO}{Radar-Inertial Odometry}
\acrodef{LIO}{LiDAR-Inertial Odometry}
\acrodef{NDT}{Normal Distribution Transform}
\acrodef{UKF}{Unscented Kalman Filter}
\acrodef{MAE}{Mean Absolute Error}
\acrodef{TTD}{Total Traveled Distance}
\acrodef{ICP}{Iterative Closest Point}
\acrodef{FoV}{Field of View}
\acrodef{FIFO}{First In First Out}
\acrodef{SVD}{Singular Value Decomposition}
\acrodef{SoC}{System-on-Chip}
\acrodef{RO}{Radar Odometry}
\acrodef{DCS}{Dynamic Covariance Scaling}
\acrodef{LSA}{Linear Sum Assignment}
\acrodef{SA}{Self-Attention}
\acrodef{mmWave}{millimeter-wave}

\usepackage{multirow,bigdelim}

\hypersetup{
    hidelinks
}

\usepackage[noabbrev,capitalise]{cleveref} 

\title{\LARGE \bf
Learning Point Correspondences In Radar 3D Point Clouds For Radar-Inertial Odometry
}

\author{Jan Michalczyk$^{1}$, Stephan Weiss$^{1}$, and Jan Steinbrener$^{1}$
\thanks{$^{1}$All authors are with the Control of Networked Systems Group, University of Klagenfurt, Austria {\tt\small \{firstname.lastname\}@ieee.org}}
\thanks{This research received funding from the Austrian Ministry of Climate Action and Energy (BMK) under the grant agreement 880057 (CARNIVAL).}%
\thanks{{\textbf{Pre-print version, accepted June/2025 (IROS), DOI follows ASAP~\copyright IEEE.}}}
}

\begin{document}

\maketitle
\thispagestyle{empty}
\pagestyle{empty}

\begin{abstract}
Using 3D point clouds in odometry estimation in robotics often requires finding a set of correspondences between points in subsequent scans. While there are established methods for point clouds of sufficient quality, state-of-the-art still struggles when this quality drops. Thus, this paper presents a novel learning-based framework for predicting robust point correspondences between pairs of \emph{noisy, sparse and unstructured 3D point clouds} from a light-weight, low-power, inexpensive, consumer-grade \ac{SoC} \ac{FMCW} radar sensor. Our network is based on the transformer architecture which allows leveraging the attention mechanism to discover pairs of points in consecutive scans with the greatest mutual affinity. The proposed network is trained in a self-supervised way using set-based multi-label classification cross-entropy loss, where the ground-truth set of matches is found by solving the \ac{LSA} optimization problem, which avoids tedious hand annotation of the training data. Additionally, posing the loss calculation as multi-label classification permits supervising on point correspondences directly instead of on odometry error, which is not feasible for sparse and noisy data from the \ac{SoC} radar we use. We evaluate our method with an open-source state-of-the-art \ac{RIO} framework in real-world \ac{UAV} flights and with the widely used public Coloradar dataset. Evaluation shows that the proposed method improves the position estimation accuracy by over \unit[14]{\%} and \unit[19]{\%} on average, respectively. The open source code and datasets can be found here: \texttt{\url{https://github.com/aau-cns/radar_transformer}}. 
\end{abstract}
\section{Introduction}\label{sec:intor}
Using radar sensors for robot localization has recently been gaining significant attention in robotics mostly owing to the advances achieved in the \ac{FMCW} \ac{mmWave} radar technology and its widespread use in the automotive industry \cite{hasch2012millimeter}. Using millimeter wavelengths makes the radar largely unaffected by air obscurants and extreme illumination, allowing it to operate in adverse conditions including snow, fog, dust or darkness \cite{morten}. Thus, combining an \ac{IMU} with a radar into \ac{RIO} offers an accurate odometry estimator robust in environments where other sensors are prone to fail \cite{ssrrjan}. Nonetheless, the light-weight and low-power \ac{SoC} variant of the \ac{mmWave} \ac{FMCW} radar which is of greatest interest \cite{radar_survey} in applications involving small-sized robots with limited payload like wheeled robots or \ac{UAV}s is known for its notoriously noisy and sparse measurements \cite{pcgen}. 

Approaches employing \ac{FMCW} \ac{SoC} radar data in \ac{RIO} estimation could generally be divided into methods relying only on the Doppler velocity information from the current measurement \cite{doer2020radar, doer2020ekf, doer_manh, kramer2020radar, kim2025, kim_2_fog} and methods performing some form of 3D point matching using past and current measurements in addition to the instantaneous Doppler velocity information \cite{jan_icra, previous_iros, almalioglu2020milli, lessismore, iriom}. Methods using bulky and expensive scanning radars for \ac{RO} often rely on keypoints extraction and scan registration owing to the high-quality and density of the 2D \unit[360]{\degree} scans they produce \cite{cen2018precise, barnes2020under, Burnett2021RadarOC}. In this work we are focused on the low-cost, lightweight and low-power \ac{FMCW} \ac{SoC} radar sensors as our primary application are small-sized \ac{UAV}s.

\begin{figure}[thpb]
  \centering
  \includegraphics[width=1.0\columnwidth]{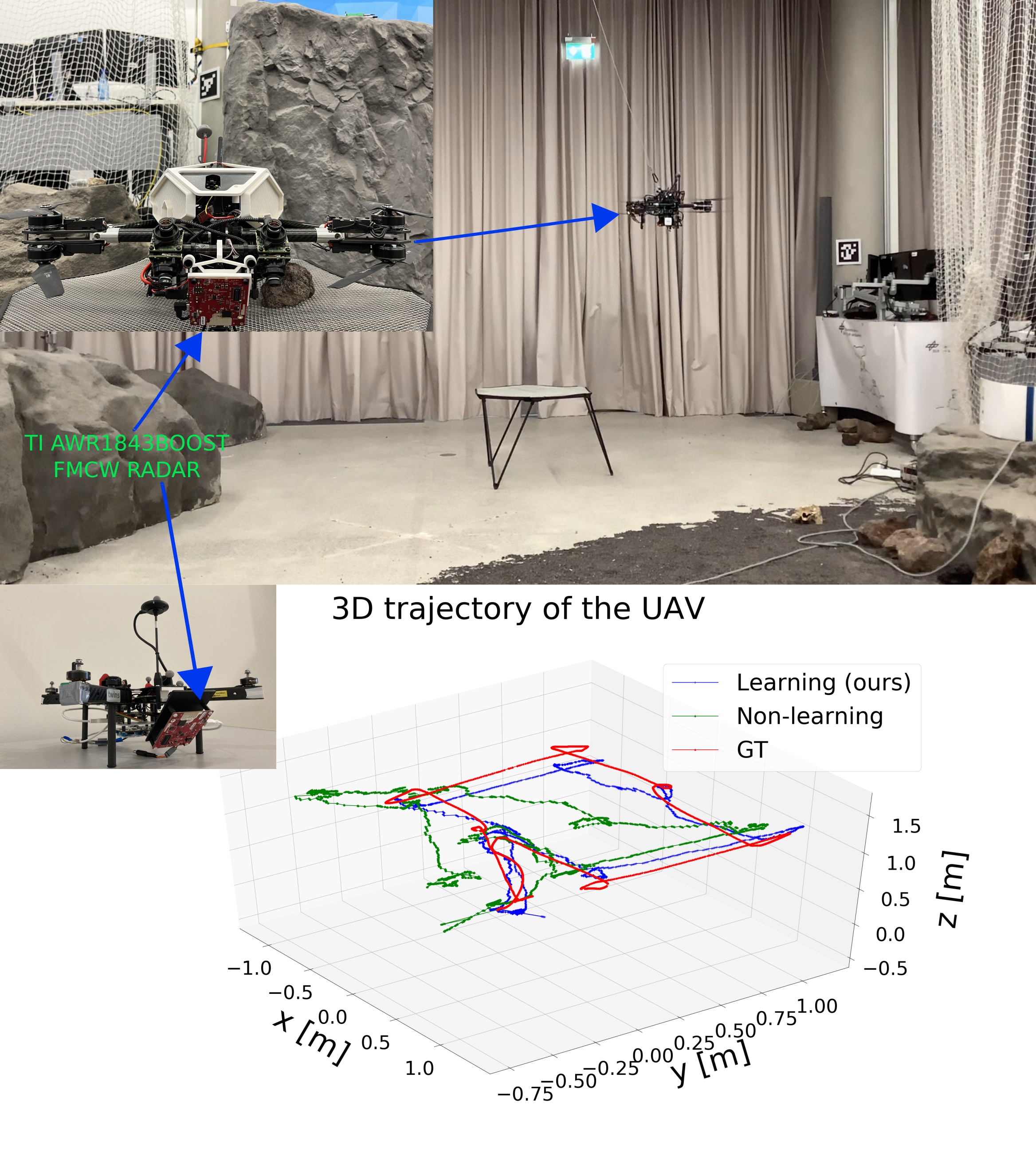}
  \caption{ARDEA-X \cite{ardea} and CNS-\ac{UAV} platforms used in this work with the mounted consumer-grade \ac{FMCW} \ac{SoC} radar sensor. The radar chip that we use outputs highly noisy and sparse 4D point clouds (3D points and Doppler velocities). In the lower part of the figure we plot the estimated position for one of the validation flights using the EKF-based \ac{RIO} framework from \cite{jan_icra} (open-sourced with \cite{jan_fg}) when switching between matching methods and using \textbf{\textit{solely}} 3D point matches in the update step. Note how the proposed method allows tracking the position of the \ac{UAV} making \textbf{\textit{no}} use of the Doppler information, while the non-learning approach drifts considerably.}
  \label{fig:platform}
\end{figure}

Using the 3D point information in state estimation usually involves finding correspondences between points in subsequent point clouds. \ac{FMCW} \ac{SoC} radar point clouds are much different from their LiDAR counterparts in that they are orders of magnitude sparser which makes it impossible to sense structures in the environment \cite{pcgen}. Moreover, they exhibit temporal and viewpoint variations, that is, point clouds of the same object measured in different time instants as well as point clouds of the same object but measured from slightly different viewpoints may differ very significantly \cite{deep_rad_det}. Additionally, they contain ghost reflections caused by speckle noise and multi-path reflections \cite{gh_det}. Hence, classic techniques developed for LiDAR data, like \ac{ICP} or \ac{NDT} cannot readily be applied to radar point clouds \cite{almalioglu2020milli}. Because of the challenging nature of the radar 3D point clouds there is a need for new methods to process them in state estimation pipelines.

In this work we present a novel learning framework for predicting robust point correspondences between \ac{FMCW} \ac{SoC} radar 3D point clouds in \ac{RIO} estimation. Our framework is inspired by recent advances in deep learning for dense 3D point clouds processing in \cite{pct, dct} and tailored for the sparse and noisy radar measurements. In particular, our framework consists of the following steps (Fig.~\ref{fig:netw}): 
\begin{enumerate}
  \item Calculate the input embeddings in a \textit{per-point} manner for each of the two consecutive input point clouds using the \textit{PointNet} architecture \cite{pointnet}.
  \item Use two transformer \cite{att} sub-networks to predict a matrix whose rows and columns correspond to the points in the first and second input point clouds, respectively, and whose entries express the degree of likelihood that the corresponding row and column form a match.
  \item Solve the \ac{LSA} optimization problem on the predicted matrix to find the set of point correspondences, where each item contains the index into the first (row) and second (column) point cloud.
  \item During training, cast the problem into a multi-label classification setting by considering the column index from the \ac{LSA} solution the class label of a point in the second point cloud, which allows leveraging the cross-entropy loss function.
  \item During inference, apply acceptance and \ac{FOV} thresholds to the set of matches found in step 3 to form the output. 
\end{enumerate}
We train and test our network on a real-world, self-collected dataset consisting of 13 manually and autonomously flown \ac{UAV} trajectories from which we select 8 for training and 5 for testing.
To make the validation more thorough and comparable, we also test our approach with the public Coloradar dataset \cite{colo}. Evaluation of our method in an open-source state-of-the-art EKF-based \ac{RIO} framework from \cite{jan_fg} (which does not use a learning-based matching algorithm to find 3D point matches) shows an increase in estimation accuracy by over \unit[14]{\%} for the self-collected dataset and by \unit[19]{\%} for the Coloradar dataset, in terms of position RMSE norm. We also note that, when deactivating the Doppler information and only keeping the 3D point matches as correction information for the IMU integration in \ac{RIO}, we note a difference in accuracy of more than \unit[70]{\%} when using our method on the self-collected dataset. To our knowledge, this is the first framework for learning point correspondences in sparse and noisy 3D point clouds as available from inexpensive SoC radar sensors. Our main contributions are:
\begin{itemize}
\item Deep learning framework for predicting robust correspondences in sparse and noisy \ac{FMCW} \ac{SoC} radar 3D point clouds.
\item Efficient, self-supervised method not requiring hand-annotated ground-truth data.
\item Formulation of the learning problem as multi-label classification which allows training on the sparse and noisy 3D point clouds yet results in unambiguous matches.
\item Evaluation of the method with real-world data (our own dataset and a public benchmark Coloradar dataset \cite{colo}) in a state-of-the-art open-source \ac{RIO} estimation framework.
\item Open-source implementation of the presented architecture together with the used dataset for the benefit of the research community.
\end{itemize}
This paper is organized as follows. Section~\ref{sec:related_work} reviews the recent related work in the domain of finding point correspondences in radar point clouds. In section~\ref{sec:net}, we describe our learning framework. Subsection~\ref{subsec:arch} outlines the architecture of the presented network. In subsection~\ref{subsec:train} we describe how training and inference are performed. In section~\ref{sec:results}, we explain the experiments (subsection~\ref{subsec:exp}) and evaluations (subsection~\ref{subsec:comparison}) conducted in order to demonstrate and validate the proposed method. Finally, we present conclusions in section~\ref{sec:conclusion}.

\section{Related Work}\label{sec:related_work}
Within the radar data association approaches employed in state estimation, we can distinguish methods suitable for the dense 2D \unit[360]{\degree} scans generated by mechanically rotating radar and 3D point clouds from \ac{SoC} radar. Within the latter, we can also differentiate between methods using industry-grade and consumer-grade sensors. With our framework, we aim at inexpensive, consumer-grade radar sensors where the quality of the sensed point clouds is considerably lower in terms of number of points and noise than of those measured using industry-grade ones. Among the methods using the industry-grade \ac{SoC} sensors, in \cite{lessismore} the authors integrate the \ac{RCS} into the nearest-neighbor search for correspondences in the euclidean space to make it more resilient against the noise. In \cite{poin_uncert}, the same authors augment the method from \cite{lessismore} by precisely modeling the uncertainty of radar measurements and incorporating it into the matching algorithm, which considerably improves the accuracy. Authors in \cite{iriom} perform distribution-to-multi-distribution geometric scan-to-submap matching by introducing spatial covariances of clusters of points. In \cite{previous_iros}, the \ac{LSA} problem is solved to construct a similarity matrix on which a search guided by a local geometric coherence is used to match subsequent 3D point clouds from a consumer-grade \ac{SoC} radar. 

Approaches developed for scanning radars differ considerably from those for their \ac{SoC} counterparts, due to the different nature of the measurements they collect. In \cite{barnes2020under}, authors use a learning approach supervised on the odometry error to find salient keypoints in the 2D radar scans. Found keypoints are matched using the cosine similarities. In \cite{cen2018precise}, a matching method is presented which leverages local coherence among points forming a scan, that is, an assumption that local geometries between points are preserved across consecutive scans. This assumption is also used in \cite{previous_iros, almalioglu2020milli}. The work in \cite{Burnett2021RadarOC} presents an unsupervised learning approach to \ac{RO} in which features from 2D radar scans are matched using a differentiable softmax matcher within their proposed network architecture.  

Interestingly, authors in \cite{kubel} argue that when industry-grade sensors are used and only odometry is needed, scan matching is no longer needed and the Doppler measurements suffice. This statement is disputed in \cite{lessismore, poin_uncert} where the authors claim the vital importance of the point matching in their \ac{RIO} system despite using automotive-grade radar. At any rate, still in many systems the price, weight and power consumption requirements mandate the use of consumer-grade radar chips as the one we use in the present paper where using point matches boosts estimation accuracy. 

\section{Learning 3D Point Correspondences In Radar 3D Point Clouds}\label{sec:net} 
We base our network architecture on the one defined in \cite{dct} for registration of dense and noiseless 3D point clouds of shapes, and adapt it to our setting of learning correspondences in variable-length, sparse and noisy \ac{SoC} radar 3D point clouds.

\begin{figure*}[thpb]
  \centering
  \includegraphics[width=2\columnwidth]{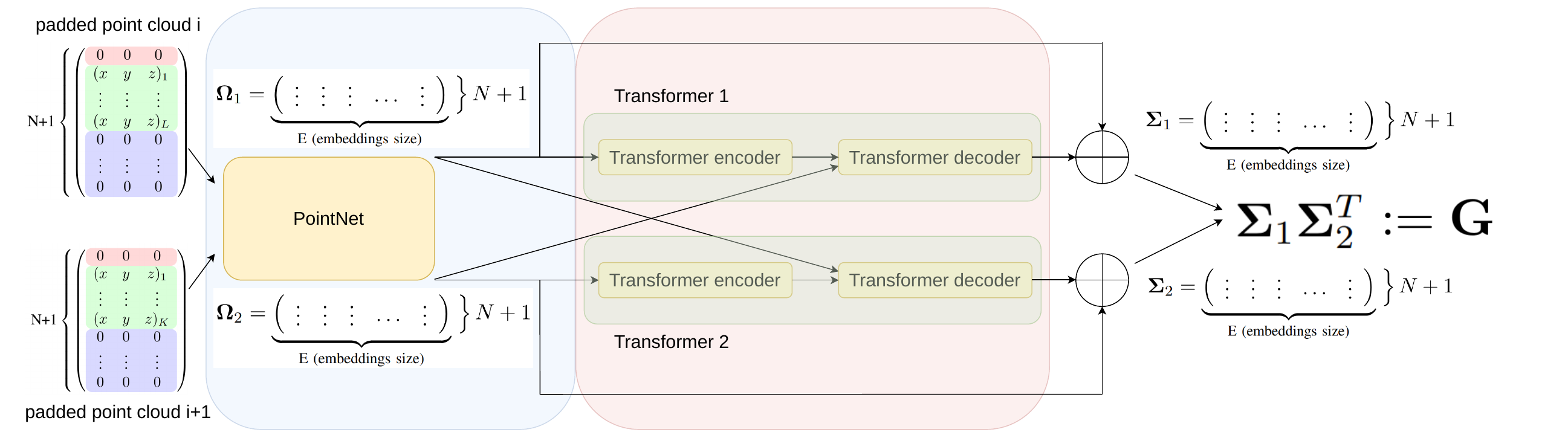}
  \caption{Our learning framework is based on the architecture proposed in \cite{dct} for dense and structured point clouds registration. We adapted it to our scenario with highly noisy and sparse consumer-grade \ac{SoC} radar 3D point clouds. Note the zero padding of the input point clouds (subsec.~\ref{subsec:train}) and the pre-pended "0", which account for the variable-length input and the class label attributed to any point with no match, respectively. Input point clouds are passed to the embedding \textit{PointNet} sub-network (light blue block). Individual embeddings enter the transformer sub-networks (light red block) where the self and the reciprocal attention is computed for each point cloud. Output matrix $\vG$ representing the mutual affinity between points in the input point clouds is obtained by calculating the dot product of the final embeddings. Within the output we search for the set of matches by solving the \ac{LSA} (see subsec.~\ref{subsec:train}).}
  \label{fig:netw}
\end{figure*}

\subsection{Network Architecture}\label{subsec:arch}
As seen in  Fig.~\ref{fig:netw}, the first step in our network applies the \textit{PointNet} sub-network to two consecutive input point clouds to embed them in a higher-dimensional space. We obtain the input point clouds by finding the length $N$ of the longest point cloud in our whole dataset and padding all point clouds to that length with zero vectors of size $1\times3$. We also append a zero vector to the beginning of each point cloud which is necessary to train our network as a multi-label classifier (see subsection~\ref{subsec:train}). That way, each of the resulting input point clouds has the shape $(N + 1)\times3$. We apply the embedding sub-network on a \textit{per-point} basis, which means that for a single point in the input represented by three coordinates $\{x, y, z\}$ we obtain an embedding vector of size $1 \times E$, where $E$ is the chosen embeddings size. \textit{PointNet} parameters are learned and shared among all points in the input. In the next step, we forward the embedded points in each point cloud to two transformer sub-networks. The role of each of the transformers is to compute new embeddings of each point cloud using the contextual information of both point clouds jointly. That way, the network can leverage the attention mechanism on both point clouds together which permits finding the embeddings tailored to the specific kind of point clouds, thus making the embeddings task-specific \cite{dct}. Specifically, in each transformer block apart from passing to the decoder the output embeddings of the encoder, we also pass in the other point cloud input embeddings. The final embeddings are calculated by summing the transformer output, which encodes the mutual information about the point clouds, with the initial embeddings. The output of the network is obtained by calculating the dot product of the final embeddings of each point in the first point cloud with the final embeddings of each point in the second point cloud. This operation yields an output matrix of shape $(N+1) \times (N+1)$. Entries in the output matrix express the affinity between points in each input point cloud, that is, the likelihood that a pair of points form a correspondence.

\subsection{Network Training And Inference}\label{subsec:train}
In the case of noisy, sparse and variable-length \ac{SoC} radar point clouds, we cannot conveniently train the network on the odometry error using the \ac{SVD} as in \cite{barnes2020under} and \cite{dct}. We thus propose a different approach to calculating the loss in our network. Namely, we treat the index of every point in the point cloud as its class label and reserve the class label (and the index) "0" for any non-matched points. The output matrix of the network is structured as follows, 
\begin{equation}\label{eq:output}
     \vG =  \underbrace{ \begin{pNiceMatrix}[margin]
      \bullet & \bullet & \cdots & \cdots & \cdots & \bullet & \bullet      \\
      \bullet & \Block[fill=green!15,rounded-corners]{4-4}{} c_{11}  & c_{12} & \cdots & c_{1K} & \cdots & \bullet \\
      \vdots  & c_{21}  & c_{22} & \cdots & c_{2K} & \cdots & \vdots \\
       \vdots & \vdots & \vdots & \ddots & \vdots & \cdots & \vdots \\
       \vdots & c_{L1} & c_{L2} & \cdots & c_{LK} & \cdots    & \vdots  \\ 
      \bullet       & \vdots   & \vdots & \vdots & \vdots &            \ddots                  & \bullet            \\
      \bullet       & \bullet              & \cdots  & \cdots & \cdots     & \bullet          & \bullet       \\
    \end{pNiceMatrix}}_{\displaystyle N+1}
    \left.\vphantom{
    \begin{pNiceMatrix}[margin]
      \bullet       & \bullet              & \cdots     & \bullet      \\
      \bullet       & \left[\begin{array}{cccc}
                            c_{11} & c_{12} & \cdots & c_{1K} \\
                             c_{21} & c_{22} & \cdots & c_{2K}\\
                             \vdots & \vdots & \ddots & \vdots \\
                             c_{L1} & c_{L2} & \cdots & c_{LK}
                       \end{array}\right]      & \cdots    & \bullet  \\ 
      \vdots       & \vdots              & \ddots                  & \bullet            \\
      \bullet       & \bullet              & \bullet               & \bullet       \\
    \end{pNiceMatrix}
    }\right\}\displaystyle{N+1}
\end{equation}
and is obtained by taking the dot product of the final embeddings $\bSigma_1$ and $\bSigma_2$ of the two consecutive input point clouds as shown in the Fig.~\ref{fig:netw}. Entries in row $i$ express the likelihood that point $i$ in the first input point cloud is a correspondence to point $j$ in the second input point cloud, where $i = 1 ... L$, $j = 1 ... K$ and $L$, $K$ are lengths of the respective point clouds. Only the green sub-matrix in the output matrix in Eq.~\ref{eq:output} carries useful information. All other elements result from adding the "0" (non-matched) class label and from padding the point clouds to equal length with zeros. In particular, appending the zero vector at the beginning of each input point cloud creates the $0$-th row and column in $\vG$. This is crucial during training, since we assign a "0" class ($0$-th index) in the ground-truth for every point in the first point cloud which does not have a match in the second point cloud.

During inference, since point clouds are usually of different lengths, we solve the \ac{LSA} problem on the green sub-matrix to find the optimal assignment,
\begin{align}
    max\sum_{i=1}^{L}\sum_{j=1}^{K}\vC_{i,j}\vX_{i,j} 
    \label{eq:linsum}
\end{align}
Where $\vX$ is a boolean matrix where $\vX_{i,j}=1$ iff row $i$ is assigned to column $j$ and $L$, $K$ are lengths of the input point clouds. $\vC$ is the green sub-matrix from Eq.~\ref{eq:output}. Constraints of the problem are such that each row is assigned to at most one column and each column to at most one row. For each entry in the solution we apply an experimentally chosen threshold to decide whether it is a match or not. \ac{LSA} is usually solved using Munkres algorithm \cite{munkres1957algorithms}.

During training, the structuring of the network output shown in Eq.~\ref{eq:output} allows us to compute the cross-entropy loss between each row (the index of which is the index of a point in the first point cloud) and the ground-truth label (index of the matched point in the second point cloud or the "0" index for a non-match), as follows,
\begin{align}\label{eq:cost}
    l_n= -\frac{1}{M}\sum_{i=1}^{M}log\left(\frac{exp(\vG(p_i, q_i))}{\sum_{j=1}^{N+1}exp(\vG(p_i, q_j))}\right) 
\end{align}
where $n = 1...B$ and $B$ is the mini-batch size, and $(p_i, q_i)$, $i = 1...M$ are the indices of the ground-truth matches in the first and second point cloud, respectively. $N+1$ is the length of each row (and column) of the $\vG$ matrix. 

Preparing the input data and ground-truth labels for training requires pre-processing. In order to generate the ground-truth point correspondences, we use the spatial transformation from the motion capture system between radar frames of every two consecutive radar measurements. Using the spatial information, we transform the 3D points from the first point cloud to the frame of the second point cloud and perform geometric matching by solving the \ac{LSA} optimization (this time minimization) problem as in Eq.~\ref{eq:linsum} but this time on a matrix whose entries are euclidean distances between points in both point clouds expressed in the second point cloud frame, as in,
\begin{align}\label{eq:scorval_C}
    \vC_{i,j}= \| \referencet{\vp}{\cR_c}{\cP_{i}}{}{c} - (\referencet{\vR}{\cR_c}{\cR_p}{}{}\referencet{\vp}{\cR_p}{\cP_{j}}{}{p}+\reference{\vp}{\cR_c}{\cR_p}{}) \| 
\end{align}
where $\reference{\vp^{\{p, c\}}}{\cR_{\{p, c\}}}{\cP}{}$ are all points from the previous radar scan at time instance $t_{p}$ and from the current radar scan at $t_{c}$, in the previous and current radar frames, respectively. $\referencet{\vR}{\cR_c}{\cR_p}{}{}$ and $\reference{\vp}{\cR_c}{\cR_p}{}$ are rotation and translation parts of the spatial transform between the current and previous radar frames obtained from the motion capture system. That way, we obtain the ground-truth class labels (indices of matched points in each point cloud). We shift the obtained labels by one to account for the "0" class for every non-matched point. Pre-processing the input radar data consists only of removing the points outside of the \ac{FOV} of the sensor and aforementioned zero padding. 
\begin{figure*}[thpb]
  \centering
  \includegraphics[width=2.0\columnwidth]{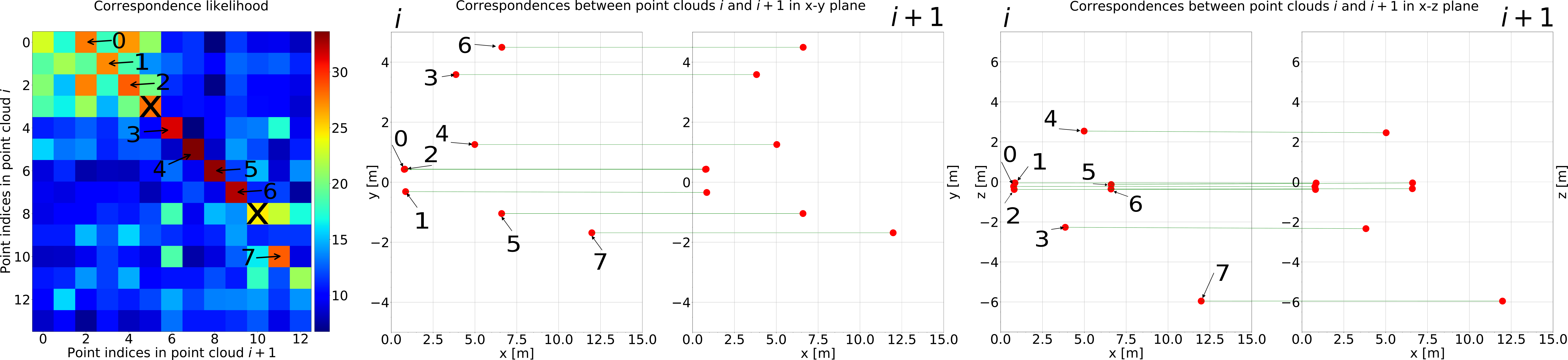}
  \caption{Leftmost part of the figure shows the correspondence likelihood matrix (the green part of the $\vG$ matrix in the Eq.~\ref{eq:output}) inferred from the learned proposed model for two consecutive point clouds $i$ and $i+1$. Rightmost and the middle parts show the resulting 3D point correspondences for the same point clouds. Correspondences are shown on $xy$ (middle sub-plot) and $xz$ (right sub-plot) planes. Numbers marking the entries in the matrix are consistent with the indices of the matches in sub-plots. During inference, we prune the matches outside the \ac{FOV} of the radar and with correspondence likelihoods below the empirically found acceptance threshold (shown as "x" in the leftmost plot). Thus, some entries in the matrix have no corresponding matches despite having relatively hot values. Only matched points are shown to not clutter the figure.}
  \label{fig:corresp}
\end{figure*}
\section{Results}\label{sec:results}

\subsection{Experiments}\label{subsec:exp}
In order to train and validate our learning framework, we collect a dataset consisting of 13 \ac{UAV} trajectories with two different platforms (ARDEA-X and CNS-UAV) described in \cite{jan_fg} (see Fig.~\ref{fig:platform}). Both platforms use the same consumer-grade TI AWR1843BOOST \ac{FMCW} \ac{SoC} radar chip mounted and configured in the same way, as well as the same pixhawk \ac{IMU} sensor. We record radar and \ac{IMU} sensor measurements as well as the ground truth pose of the \ac{UAV} using a motion capture system. We divide this dataset into 8 training and 5 validation trajectories. The training dataset contains trajectories between \unit[150]{m} - \unit[180]{m} in length flown manually. The validation dataset contains shorter trajectories between \unit[11]{m} - \unit[38]{m}, among which some are manually flown while others are pre-planned, executed using specified waypoints. We also validate our method on five sequences from the public open-source Coloradar dataset \cite{colo}. Coloradar sequences are collected using hand-held sensor rig containing the same TI radar chip as ARDEA-X and CNS-UAV platforms. Coloradar sequences contain much more aggressive motion than the self-collected dataset, thus being more challenging. While in three sequences a motion capture system is used as the ground-truth, the other two contain high-precision \ac{LIO} data. We train our network using PyTorch open-source package. We assess our 3D point matching framework qualitatively in an indirect way by plugging it into an open-source \ac{RIO} framework from \cite{jan_icra, jan_fg} and comparing the accuracy of the obtained estimates to the case when the default (non-learning) matching algorithm is used. Between executions of the \ac{RIO} estimator, we only exchange the matching algorithm, all other parameters and settings remain the same. The \ac{RIO} which we use for validation is EKF-based and in the update step uses three sources of information: 3D point matches, Doppler velocities and persistent features. For the self-collected dataset, we execute the \ac{RIO} in two configurations: in the default configuration with both Doppler and point matches, and with only point matches enabled in the update step. For the Coloradar we only use default configuration (point matches and Doppler).
In all cases, we compile the \ac{RIO} framework without persistent features in the update. In the case of our learning-based framework, we execute the inference on the learned model in a Python node before feeding it to the \ac{RIO}. The inference with a non-optimized model takes on average \unit[0.0273]{s}, which means the optimized implementation would lend itself to real-time use. The non-learning matching algorithm is implemented within the \ac{RIO} estimator as its default matching algorithm and described in \cite{previous_iros}. Both \ac{RIO} and inference node are executed offline on the recorded sensor data on an Intel Core i7-10850H vPRO laptop with 16 GB RAM.

\begin{figure*}[thpb]
  \centering
  \includegraphics[width=2.0\columnwidth]{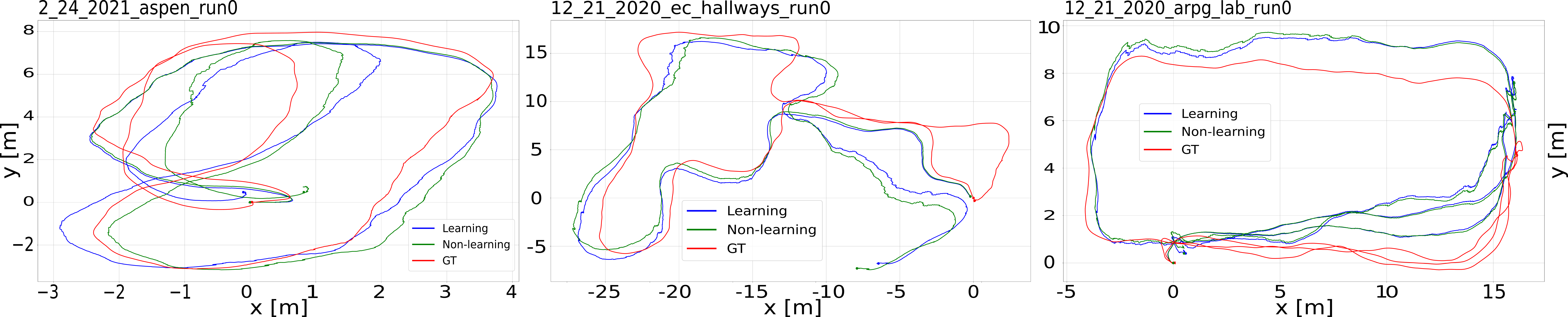}
  \caption{Three illustrative sequences (also used in \cite{poin_uncert}) out of five chosen from the Coloradar dataset for evaluation. From left to right: "2\_24\_2021\_aspen\_run0", "12\_21\_2020\_ec\_hallways\_run0", "12\_21\_2020\_arpg\_lab\_run0". In \textcolor{red}{red} we mark the ground-truth, and in \textcolor{green}{green} the non-learning and in \textcolor{blue}{blue} the learning (proposed) approaches, respectively. Across all used Coloradar sequences, using our learning-based matching method results in a decrease of the norm of position RMSE by \unit[19]{\%}.}
  \label{fig:colo2d}
\end{figure*}

\begin{figure}[thpb]
  \centering
  \includegraphics[width=1.\columnwidth]{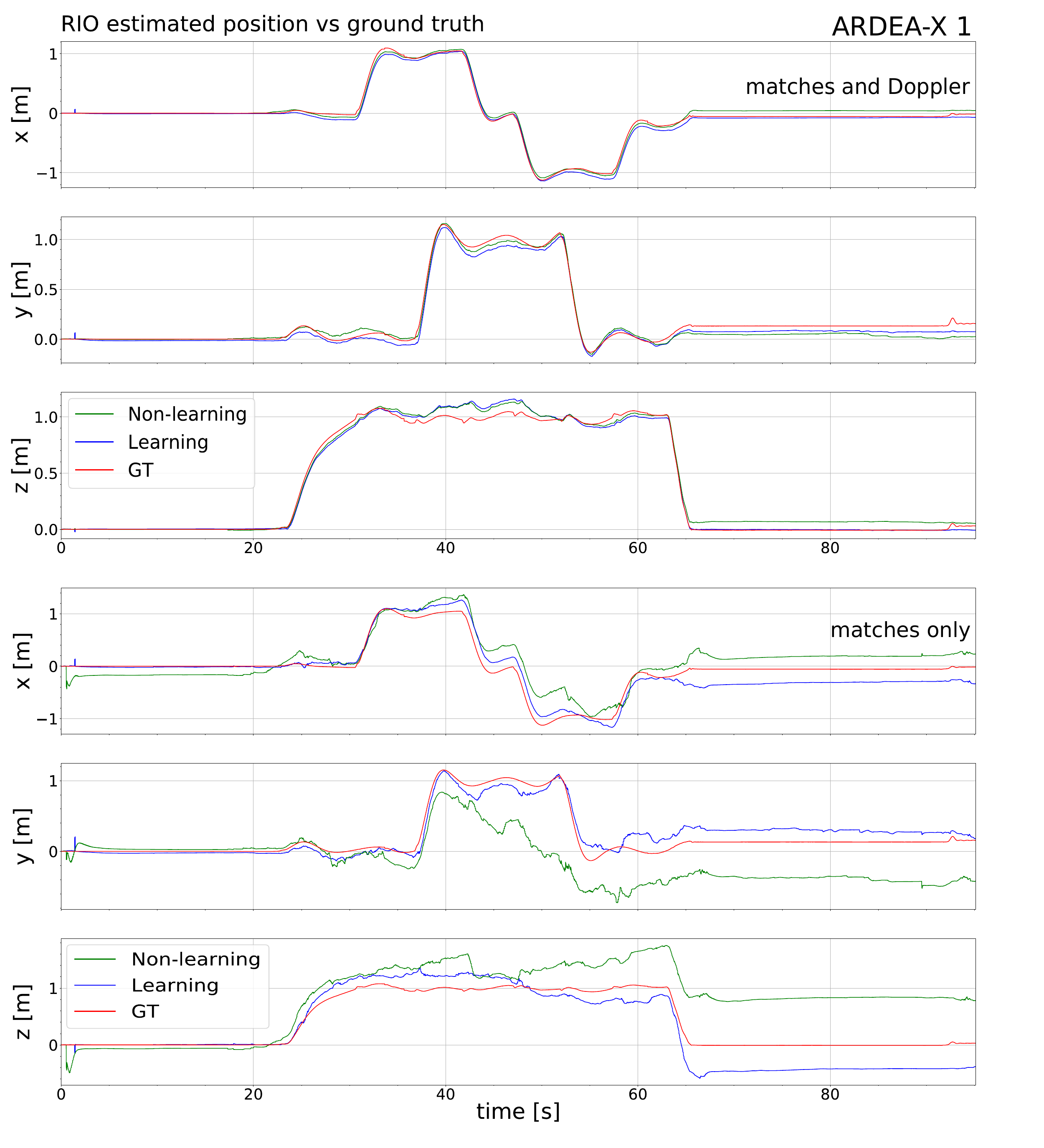}
  \caption{Estimated position of the ARDEA-X \ac{UAV} for the flight $1$ from the Tab.~\ref{tab:rmses_att} and \ref{tab:rmses}. We plot the $x, y, z$ coordinates of the estimate against the ground-truth for the configuration with matches and Doppler, and only matches used in the update step of the EKF \ac{RIO} framework used for validation. Each configuration is executed with the proposed learning-based and the default non-learning matching algorithm. In \textcolor{red}{red} the ground-truth, in \textcolor{green}{green} and \textcolor{blue}{blue} non-learning and learning approaches, respectively.}
  \label{fig:2dpos}
\end{figure}

\begin{figure}[thpb]
  \centering
  \includegraphics[width=1.\columnwidth]{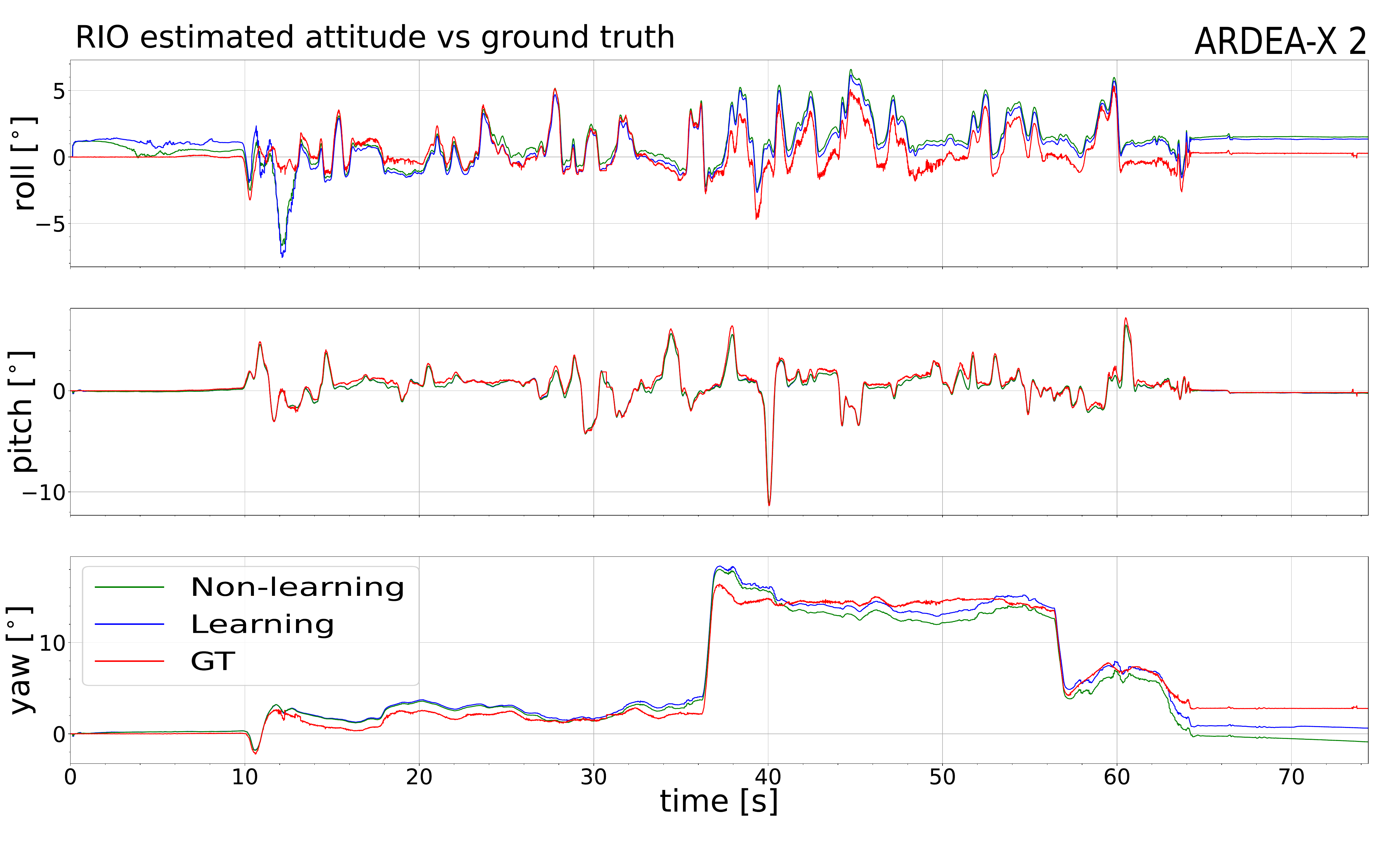}
  \caption{Estimated attitude of the ARDEA-X \ac{UAV} for the flight $2$ from the Tab.~\ref{tab:rmses_att} and \ref{tab:rmses}. We plot the roll, pitch and yaw angles of the estimate against the ground-truth for the configuration with matches and Doppler used in the update step of the EKF in \ac{RIO} framework used for validation, when executed with the proposed learning-based (in \textcolor{blue}{blue}) and the default non-learning matching algorithm (in \textcolor{green}{green}). In \textcolor{red}{red} we plot the ground-truth attitude.}
  \label{fig:2datt}
\end{figure}

\subsection{Evaluation}\label{subsec:comparison}
For each trajectory from both validation datasets (self-collected and Coloradar), we compute the norm of RMSE of the position and attitude estimates along with the mean and the standard deviation of the obtained values when switching the matching algorithm inside the \ac{RIO} estimator (see Tab.~\ref{tab:rmses} and Tab.~\ref{tab:rmses_att}). In the case of the self-collected dataset, we compute the norm of RMSE values in the case when both point matches and Doppler are used, and additionally, when only point matches residuals are used in the update step of the EKF in the \ac{RIO}. Our comparisons show that when only point matches are used in the update step, which is the most direct way of assessing the performance of our learning-based matching framework, we obtain a striking \unit[70.38]{\%} improvement in the position estimate accuracy on average. When compared to the state-of-the-art configuration of the \ac{RIO}, that is, with both Doppler and point matches residuals enabled, we obtain a \unit[14.28]{\%} improvement in position accuracy on average. With the Coloradar dataset, we only execute the full configuration containing both Doppler and point matches residuals and obtain \unit[19.01]{\%} improvement in position accuracy on average. The motion in the Coloradar dataset is too aggressive for the configuration using only point matches to work properly (for both learning and non-learning). For our recorded dataset, as can be seen in Fig.~\ref{fig:2dpos}, when point matches are used as the sole source of information for measurement updates in the EKF \ac{RIO}, the presented method allows for much more accurate estimation than the non-learning approach and when combined with the Doppler velocity measurements, greatly reduces the final error.

In Fig.~\ref{fig:corresp}, we can see how the trained network infers the correspondences. Note how matches $0$ and $2$ lie very close geometrically and hence points involved in them have all high mutual affinities, nevertheless, the network still makes correct distinction between them. Points involved in match $1$, which also lie close to points involved in matches $0$ and $2$, do not have high affinity with points in $0$, $2$. This can be explained by looking at the the middle plot and observing that on the $xy$-plane, points in match $1$ are offset from points in $0$ and $2$. For points in matches $5$ and $6$, despite all of them being close on the $xz$-plane, our network correctly assigns the mutual affinities, since on the $xy$-plane, the points are clearly separated resulting in unambiguous matches. Points in matches $3, 4, 7$ are significantly spaced in both planes, thus all have strong unambiguous mutual affinity values. Points $(3, 5)$ are not considered a match despite their high mutual affinity because at least one of them is outside of the \ac{FOV}. Similarly, the points $(8, 10)$ are not counted as a match since their correspondence likelihood value is below the empirically determined acceptance threshold.

In Fig.~\ref{fig:colo2d} we plot the estimation results for three out of chosen five Coloradar sequences for both used matching methods. From the five sequences, "12\_21\_2020\_ec\_hallways\_run0", "2\_24\_2021\_aspen\_run0" and "12\_21\_2020\_arpg\_lab\_run0" are also chosen in the latest state-of-the-art work on \ac{RIO} presented in \cite{poin_uncert} where the authors also provide the norm of RMSE of position and attitude estimate for their method. This allows us to note that for sequences "2\_24\_2021\_aspen\_run0" and "12\_21\_2020\_arpg\_lab\_run0" our method outperforms \cite{poin_uncert} $\unit[3.044]{m}$ (ours) to $\unit[3.820]{m}$ and $\unit[5.388]{m}$ (ours) to $\unit[6.101]{m}$, respectively in position and $\unit[13.372]{\degree}$ (ours) to $\unit[30.905]{\degree}$, and $\unit[8.287]{\degree}$ (ours) to $\unit[12.640]{\degree}$, respectively in attitude. For the sequence "12\_21\_2020\_ec\_hallways\_run0" our method performs worse, $\unit[9.927]{m}$ (ours) to $\unit[5.223]{m}$ in position as well as in attitude $\unit[18.357 ]{\degree}$ (ours) to $\unit[16.070]{\degree}$. 

Note, however, that in \cite{poin_uncert} the underlying estimator is different to the one we use here. Thus, we compare in Tab.~\ref{tab:rmses} and Tab.~\ref{tab:rmses_att} more rigorously the specific benefit of the learning-based matching.
For this, we implement the \ac{RIO} framework of~\cite{jan_icra, jan_fg} with Doppler and point matches as well as only point matches from our learning based approach and with the original non-learning approach. Tab.~\ref{tab:rmses} clearly shows that our learning based approach improves the performance in the position estimation in almost all runs. When using both Doppler and point matches, the improvement using our approach is in average over \unit[14]{\%} or a bit over \unit[5]{cm} on our own datasets, and \unit[19]{\%} or \unit[8]{cm} on the Coloradar datasets. With a standard deviation much higher than the improvements, these results have limited statistical relevance. \emph{However, when using only point matches, the improvement is much higher with over \unit[70]{\%} or nearly \unit[1.5]{m} on our datasets. With a standard deviation of a bit over \unit[18]{cm}, this clearly underlines the estimation improvement due to our approach}. The Coloradar dataset trajectories are too agile for the estimator to work properly when not including Doppler information, hence the 'x' in the lower right part of the table.

The benefit of our approach regarding the attitude estimation is less clear. Using both Doppler and point matches we observe in average a decrease in performance of \unit[5]{\%} or roughly half a degree on both our and the Coloradar datasets. When only using point matches, we observe in average a bit more than \unit[9]{\%} or nearly 1 degree performance drop on our datasets. Note, however, that these differences are barely statistically relevant since the standard deviation in all cases is higher than 5 degrees for our approach. Thus, we can conclude that while our method has a clearly positive impact on the position estimate, the attitude barely benefits from the new approach. The root cause of the reduced benefit in attitude is to be investigated further -- we assume a connection to the bad angular resolution and high angular noise that comes with this type of sensors.
%
%
We plot the attitude estimates for one of the trajectories from the self-collected dataset in the Fig.~\ref{fig:2datt}. 

\begin{table}[]
\centering
\caption{RMSE norm values of position estimate for both matching methods across self-collected \ac{UAV} flights and sequences from open-source Coloradar dataset.}
\label{tab:rmses}
\begin{tabular}{|c|cccc|}
\hline
\multirow{3}{*}{Nr} & \multicolumn{4}{c|}{\begin{tabular}[c]{@{}c@{}}Self-collected dataset\\ $\|$RMSE$\|$ of position {[}m{]}\end{tabular}} \\ \cline{2-5} 
 & \multicolumn{2}{c|}{Doppler and matches} & \multicolumn{2}{c|}{Matches only} \\ \cline{2-5} 
 & \multicolumn{1}{c|}{\begin{tabular}[c]{@{}c@{}}Learning\\ (ours)\end{tabular}} & \multicolumn{1}{c|}{\begin{tabular}[c]{@{}c@{}}Non\\ learning\end{tabular}} & \multicolumn{1}{c|}{\begin{tabular}[c]{@{}c@{}}Learning\\ (ours)\end{tabular}} & \begin{tabular}[c]{@{}c@{}}Non\\ learning\end{tabular} \\ \hline
1 & \multicolumn{1}{c|}{\textbf{0.083}} & \multicolumn{1}{c|}{0.101} & \multicolumn{1}{c|}{\textbf{0.351}} & 0.748 \\ \hline
2 & \multicolumn{1}{c|}{\textbf{0.212}} & \multicolumn{1}{c|}{0.272} & \multicolumn{1}{c|}{\textbf{0.793}} & 1.795 \\ \hline
3 & \multicolumn{1}{c|}{0.714} & \multicolumn{1}{c|}{\textbf{0.704}} & \multicolumn{1}{c|}{\textbf{0.745}} & 1.608 \\ \hline
4 & \multicolumn{1}{c|}{0.380} & \multicolumn{1}{c|}{\textbf{0.338}} & \multicolumn{1}{c|}{\textbf{0.431}} & 1.074 \\ \hline
5 & \multicolumn{1}{c|}{\textbf{0.230}} & \multicolumn{1}{c|}{0.473} & \multicolumn{1}{c|}{\textbf{0.736}} & 5.088 \\ \hline
Average & \multicolumn{1}{c|}{0.324} & \multicolumn{1}{c|}{0.378} & \multicolumn{1}{c|}{0.611} & 2.063 \\ \hline
Std. dev. & \multicolumn{1}{c|}{0.217} & \multicolumn{1}{c|}{0.202} & \multicolumn{1}{c|}{0.183} & 1.558 \\ \hline
Sequence & \multicolumn{4}{c|}{\begin{tabular}[c]{@{}c@{}}Coloradar dataset\\ $\|$RMSE$\|$ of position {[}m{]}\end{tabular}} \\ \hline
aspen\_run0 & \multicolumn{1}{c|}{\textbf{3.044}} & \multicolumn{1}{c|}{5.327} & \multicolumn{1}{c|}{x} & x \\ \hline
arpg\_lab\_run0 & \multicolumn{1}{c|}{\textbf{5.388}} & \multicolumn{1}{c|}{6.080} & \multicolumn{1}{c|}{x} & x \\ \hline
\multicolumn{1}{|l|}{ec\_hallways\_run0} & \multicolumn{1}{c|}{\textbf{9.927}} & \multicolumn{1}{c|}{11.523} & \multicolumn{1}{c|}{x} & x \\ \hline
aspen\_run4 & \multicolumn{1}{c|}{\textbf{5.315}} & \multicolumn{1}{c|}{6.296} & \multicolumn{1}{c|}{x} & x \\ \hline
aspen\_run5 & \multicolumn{1}{c|}{\textbf{3.466}} & \multicolumn{1}{c|}{4.280} & \multicolumn{1}{c|}{x} & x \\ \hline
Average & \multicolumn{1}{c|}{5.428} & \multicolumn{1}{c|}{6.701} & \multicolumn{1}{c|}{x} & x \\ \hline
Std. dev. & \multicolumn{1}{c|}{2.728} & \multicolumn{1}{c|}{2.808} & \multicolumn{1}{c|}{x} & x \\ \hline
\end{tabular}
\end{table}

\begin{table}[]
\centering
\caption{RMSE norm values of attitude estimate for both matching methods across self-collected \ac{UAV} flights and sequences from open-source Coloradar dataset.}
\label{tab:rmses_att}
\begin{tabular}{ccccc}
\hline
\multicolumn{1}{|c|}{\multirow{3}{*}{Nr}} & \multicolumn{4}{c|}{\begin{tabular}[c]{@{}c@{}}Self-collected dataset\\ $\|$RMSE$\|$ of attitude {[}\degree{]}\end{tabular}} \\ \cline{2-5} 
\multicolumn{1}{|c|}{} & \multicolumn{2}{c|}{Matches and Doppler} & \multicolumn{2}{c|}{Matches only} \\ \cline{2-5} 
\multicolumn{1}{|c|}{} & \multicolumn{1}{c|}{\begin{tabular}[c]{@{}c@{}}Learning\\ (ours)\end{tabular}} & \multicolumn{1}{c|}{\begin{tabular}[c]{@{}c@{}}Non\\ learning\end{tabular}} & \multicolumn{1}{c|}{\begin{tabular}[c]{@{}c@{}}Learning\\ (ours)\end{tabular}} & \multicolumn{1}{c|}{\begin{tabular}[c]{@{}c@{}}Non\\ learning\end{tabular}} \\ \hline
\multicolumn{1}{|c|}{1} & \multicolumn{1}{c|}{3.996} & \multicolumn{1}{c|}{\textbf{3.203}} & \multicolumn{1}{c|}{\textbf{0.966}} & \multicolumn{1}{c|}{6.689} \\ \hline
\multicolumn{1}{|c|}{2} & \multicolumn{1}{c|}{\textbf{1.642}} & \multicolumn{1}{c|}{2.036} & \multicolumn{1}{c|}{4.730} & \multicolumn{1}{c|}{\textbf{3.597}} \\ \hline
\multicolumn{1}{|c|}{3} & \multicolumn{1}{c|}{17.068} & \multicolumn{1}{c|}{\textbf{16.945}} & \multicolumn{1}{c|}{17.198} & \multicolumn{1}{c|}{\textbf{17.067}} \\ \hline
\multicolumn{1}{|c|}{4} & \multicolumn{1}{c|}{10.058} & \multicolumn{1}{c|}{\textbf{9.712}} & \multicolumn{1}{c|}{8.547} & \multicolumn{1}{c|}{\textbf{6.234}} \\ \hline
\multicolumn{1}{|c|}{5} & \multicolumn{1}{c|}{18.158} & \multicolumn{1}{c|}{\textbf{16.551}} & \multicolumn{1}{c|}{21.108} & \multicolumn{1}{c|}{\textbf{14.305}} \\ \hline
\multicolumn{1}{|c|}{Average} & \multicolumn{1}{c|}{10.184} & \multicolumn{1}{c|}{9.689} & \multicolumn{1}{c|}{10.510} & \multicolumn{1}{c|}{9.578} \\ \hline
\multicolumn{1}{|c|}{Std. dev.} & \multicolumn{1}{c|}{7.453} & \multicolumn{1}{c|}{7.077} & \multicolumn{1}{c|}{8.446} & \multicolumn{1}{c|}{5.782} \\ \hline
\multicolumn{1}{|c|}{Sequence} & \multicolumn{4}{c|}{\begin{tabular}[c]{@{}c@{}}Coloradar dataset\\ $\|$RMSE$\|$ of attitude {[}\degree{]}\end{tabular}} \\ \hline
\multicolumn{1}{|c|}{aspen\_run0} & \multicolumn{1}{c|}{13.372} & \multicolumn{1}{c|}{\textbf{9.389}} & \multicolumn{1}{c|}{x} & \multicolumn{1}{c|}{x} \\ \hline
\multicolumn{1}{|c|}{arpg\_lab\_run0} & \multicolumn{1}{c|}{8.287} & \multicolumn{1}{c|}{\textbf{7.506}} & \multicolumn{1}{c|}{x} & \multicolumn{1}{c|}{x} \\ \hline
\multicolumn{1}{|c|}{ec\_hallways\_run0} & \multicolumn{1}{c|}{\textbf{18.357}} & \multicolumn{1}{c|}{21.666} & \multicolumn{1}{c|}{x} & \multicolumn{1}{c|}{x} \\ \hline
\multicolumn{1}{|c|}{aspen\_run4} & \multicolumn{1}{c|}{\textbf{7.776}} & \multicolumn{1}{c|}{8.117} & \multicolumn{1}{c|}{x} & \multicolumn{1}{c|}{x} \\ \hline
\multicolumn{1}{|c|}{aspen\_run5} & \multicolumn{1}{c|}{9.363} & \multicolumn{1}{c|}{\textbf{7.947}} & \multicolumn{1}{c|}{x} & \multicolumn{1}{c|}{x} \\ \hline
\multicolumn{1}{|c|}{Average} & \multicolumn{1}{c|}{11.431} & \multicolumn{1}{c|}{10.925} & \multicolumn{1}{c|}{x} & \multicolumn{1}{c|}{x} \\ \hline
\multicolumn{1}{|c|}{Std. dev.} & \multicolumn{1}{c|}{6.045} & \multicolumn{1}{c|}{4.451} & \multicolumn{1}{c|}{x} & \multicolumn{1}{c|}{x} \\ \hline
\multicolumn{1}{l}{} & \multicolumn{1}{l}{} & \multicolumn{1}{l}{} & \multicolumn{1}{l}{} & \multicolumn{1}{l}{} \\
\multicolumn{1}{l}{} & \multicolumn{1}{l}{} & \multicolumn{1}{l}{} & \multicolumn{1}{l}{} & \multicolumn{1}{l}{}
\end{tabular}
\end{table}

\section{Conclusions}\label{sec:conclusion}
In this paper, we presented a novel self-supervised learning framework for finding 3D point correspondences in sparse and noisy point clouds from a \ac{SoC} \ac{FMCW} radar. To our knowledge, this is the first learning approach addressing the problem of data association in the challenging setting of 3D point cloud measurements from low-cost, low-power, lightweight, consumer-grade radar chips. In our framework, we leverage the \textit{PointNet} architecture to compute individual point embeddings in each of the two consecutive input point clouds. Subsequently, using transformer architecture and its attention mechanism, we augment the initial embeddings with the reciprocal information from both inputs, to finally form the matching likelihood matrix by calculating the dot product of the augmented embeddings. We provide a self-supervision method using set-based multi-label classification cross-entropy loss, where the ground-truth set of matches is calculated by solving the \ac{LSA} optimization problem. Employing multi-label classification cross-entropy loss enables directly using  correspondences in training. This is crucial since training on odometry error using e.g. \ac{SVD}, as used in methods for scanning radars or dense point clouds, is not feasible with the sparse and noisy measurements from the \ac{SoC} radar sensor that we use in this work. We applied our framework to the task of \ac{RIO} estimation on a small-sized \ac{UAV} and showed that it outperforms the default non-learning 3D point matching method. In particular, in an open-source state-of-the-art \ac{RIO} framework we switched the 3D point matching algorithm from the default non-learning one to the one presented in this paper while keeping all other settings and parameters unchanged. The reduction in the norm of RMSE of position estimate calculated over the whole real-world validation dataset in both cases when only matches, and matches with Doppler velocity are used in the estimator reveals that our learning-based method surpasses the non-learning one. We make both our framework and the datasets open-source for the benefit of the research community.

\section*{Acknowledgment}
Authors would like to thank Florian Steidle, Julius Quell and Marcus G. M\"uller from the MAV Exploration Team at the Institute of Robotics and Mechatronics in the German Aerospace Center (DLR) for hosting the author and helping with the dataset acquisition.


\balance
\bibliographystyle{IEEEtran}
\bibliography{root.bib}

\end{document}